\documentclass[twoside]{article}

\usepackage[accepted]{aistats2022}

\usepackage[dvipsnames]{xcolor}

\usepackage{hyperref}
\usepackage{url}
\usepackage{physics}
\usepackage{amsmath, amsthm, amssymb, amsfonts}
\usepackage{graphicx}
\usepackage[authoryear]{natbib}
\usepackage{bbm}
\usepackage{cleveref}

\def\approachName{GalilAI}
\newtheorem{definition}{Definition}
\DeclareMathOperator*{\argmin}{arg\,min}
\DeclareMathOperator*{\argmax}{arg\,max}
\newcommand{\powerset}{\raisebox{.15\baselineskip}{\Large\ensuremath{\wp}}}

\newcommand{\testEnv}[0]{\ensuremath{\vb{p}'}}
\newcommand{\trainEnv}[0]{\ensuremath{\vb{p}}}

\begin{document}

%

%

\twocolumn[

\aistatstitle{GalilAI: Out-of-Task Distribution Detection using Causal Active Experimentation for Safe Transfer RL}

\aistatsauthor{ Sumedh A Sontakke$^*$ \And Stephen Iota$^*$ \And  Zizhao Hu$^*$ }
\aistatsaddress{ University of Southern California \And  University of Southern California \And University of Southern California } 
\aistatsauthor{ Arash Mehrjou \And Laurent Itti \And Bernhard Sch{\"o}lkopf}
\aistatsaddress{ MPI for Intelligent Systems \And  University of Southern California \And MPI for Intelligent Systems } 
]
\def\thefootnote{*}\footnotetext{These authors contributed equally to this work}

\begin{abstract}
Out-of-distribution (OOD) detection is a well-studied topic in supervised learning. Extending the successes in supervised learning methods to the reinforcement learning (RL) setting, however, is difficult due to the data generating process --- RL agents actively query their environment for data, and the data are a function of the policy followed by the agent. An agent could thus neglect a shift in the environment if its policy did not lead it to explore the aspect of the environment that shifted. Therefore, to achieve safe and robust generalization in RL, there exists an unmet need for OOD detection through active experimentation. Here, we attempt to bridge this lacuna by first defining a causal framework for OOD scenarios or environments encountered by RL agents in the wild. Then, we propose a novel task: that of Out-of-Task Distribution (OOTD) detection. We introduce an RL agent that actively experiments in a test environment and subsequently concludes whether it is OOTD or not. We name our method \approachName, in honor of Galileo Galilei, as it discovers, among other causal processes, that gravitational acceleration is independent of the mass of a body. Finally, we propose a simple probabilistic neural network baseline for comparison, which extends extant Model-Based RL. We find that \approachName~outperforms the baseline significantly. See visualizations of our method \href{https://galil-ai.github.io/}{\textcolor{blue}{here.}} 


\end{abstract}
\section{Introduction and Related Work}

\textbf{Generalization} to near-distribution shifts caused by natural perturbations and \textbf{Detection} of out-of-distribution shifts caused by artificial perturbations (adversarial attacks) are central desiderata of modern decision-making systems. 
Significant advances have been made in supervised learning systems on both fronts - with work in transfer/meta-learning aiding the ability of ML systems to generalize across shifts in input distributions \citep{schmidhuber2007godel, santoro2016meta, finn2017model}. Such methods learn internal representations which are invariant to perturbations occurring in data \citep{bengio2013deep}. These invariant representations are subsequently used for domain adaptation \citep{zhao2019learning,MuaBalSch13}, with applications in music \citep{blumensath2005sparse} and speech \citep{serdyuk2016invariant}. Out-of-distribution Detection for the supervised learning domain has also made significant advances \citep{hendrycks2016baseline, devries2018learning,liang2017enhancing, goodfellow2014explaining}, with the development of both training-time methods \citep{xiao2020likelihood} (alterations to typical supervised training to make models robust to OOD inputs) and inference-time methods (utilizing the features of a fully trained model to detect OOD samples)\citep{hsu2020generalized}. 

While attempts have been made in generalization in the space of sequential long-horizon RL and decision-making \citep{finn2017model, nagabandi2018learning, gupta2018meta, parisotto2015actor, rakelly2019efficient, zintgraf2019varibad}, Out-of-Distribution Detection is fairly unexplored. To our knowledge, our work is the first that offers a concrete causal framework for OOD Detection. 

We motivate the need for OOD Detection in RL with an example. Consider an agent that has learnt to land an aircraft for various values of directions and velocity of crosswinds. Now consider the situation when one of the airplane's engines fails when the agent is deployed. Current RL systems would assume that the observations they receive from this test environment were caused by perhaps high crosswinds and would subsequently increase fuel flow to the engines - a potentially disastrous strategy. On the contrary, a seasoned pilot might perform an experiment - perhaps yawing the aircraft from side-to-side, concluding that due to the low controllability of the aircraft, the engine was somehow compromised. Our work extends that of \citeauthor{sontakke2021causal} by utilizing advances in algorithmic information theory and curiosity-based reinforcement learning to ``encourage" the RL agent to perform such experimental behaviors and conclude whether a test-time environment is out-of-training-distribution or not. 

During our experiments, we find that our agent discovers the \textbf{Galilean Equivalence Principle}, managing to successfully decouple the effect of mass and gravitational acceleration. For this reason, we refer to the agent as GalilAI (pronounced Galilei).
The contributions of our work are as follows:
\begin{itemize}
    \item\textbf{Causal transfer:} We offer a causal perspective on transfer learning in RL and provide a theoretical framework for defining various classes of transfer RL problems.    
    \item \textbf{Causal active experimentation (\approachName) for safe transfer RL:} We extend the work of \citeauthor{sontakke2021causal} to provide an algorithm aimed at improving the safety of transfer reinforcement learning by detecting whether a given test environment is out-of-distribution or not. If an environment is detected as OOD, the agent could relinquish control of a system to a human operator \citep{amodei2016concrete}.    
    \item \textbf{Probabilistic baseline:} Due to a lack of prior work in the field, we propose a simple probabilistic neural network baseline for OOD Detection of environments in RL. We compare \approachName~and the PNN in complex robotic domains such as the Causal World \citep{ahmed2020causalworld} and Mujoco \citep{6386109}.
\end{itemize}
\section{Preliminaries}
\begin{definition}[Causal factors]
  \label{def:causal_factors}
  Consider the POMDP ($\mathcal{O}$, $\mathcal{S}$, $\mathcal{A}$, $\phi$, $\theta$, r) with observation space $\mathcal{O}$, state space $\mathcal{S}$, action space $\mathcal{A}$, the transition function $\phi$, emission function $\theta$, and the reward function $r$. Let  $\vb{o}_{0:T}\in \mathcal{O}^T$ denote a trajectory of observations of length $T$. Let $d(\cdot, \cdot):\mathcal{O}^T \times \mathcal{O}^T\to \mathbb{R}_+$ be a distance function defined on the space of trajectories of length $T$. The set $H=\{\vb{h}_0,\vb{h}_1, \ldots, \vb{h}_{K-1}\}$ is called a set of $\epsilon-$causal factors if for every $\vb{h}_j\in H$, there exists a unique sequence of actions $\vb{a}_{0:T}$ that clusters the observation trajectories into $m$ disjoint sets $C_{1:m}$ such that $\forall C_a, C_b$, a minimum separation distance of $\epsilon$ is ensured:
\begin{equation}
  \min \{d(\vb{o}_{0:T}, \vb{o}'_{0:T}): \vb{o}_{0:T}\in C_a, \vb{o}'_{0:T}\in C_b\} > \epsilon
\end{equation}
and that $\vb{h}_j$ is the cause of the obtained trajectory of states i.e. $\forall v \neq v'$,
\begin{equation}
  p(\vb{o}_{0:T}|do(\vb{h}_j = v), \vb{a}_{0:T}) \neq p(\vb{o}_{0:T}|do(\vb{h}_j = v'), \vb{a}_{0:T})
\end{equation}
where $do(\vb{h}_j)$ corresponds to an intervention on the value of the causal factor $\vb{h}_j$.


\end{definition}

According to \Cref{def:causal_factors}, a causal factor $\vb{h}_j$ is a variable in the environment the value of which, when intervened on (i.e., varied) using $do(\vb{h}_j)$ over a set of values, results in trajectories of observations that are divisible into disjoint clusters $C_{1:m}$ under a particular sequence of actions $\vb{a}_{0:T}$. These clusters represent the quantized values of the causal factor. For example, mass, which is a causal factor of a body, under an action sequence of a grasping and lifting motion with fixed force, may result in 2 clusters, liftable (low mass) and not-liftable (high mass).  
\subsection{POMDPs and Causal POMDPs}
\textbf{Classical POMDPs} ($\mathcal{O}$, $\mathcal{S}$, $\mathcal{A}$, $\phi$, $\theta$, r) consist of an observation space $\mathcal{O}$, state space $\mathcal{S}$, action space $\mathcal{A}$, the transition function $\phi$, emission function $\theta$, and the reward function $r$. An agent in an unobserved state $\vb{s}_t$ takes an action $\vb{a}_t$ and consequently causes a transition in the environment through $\phi(\vb{s}_{t+1}|\vb{s}_t, \vb{a}_t)$. The agent receives an observation $\vb{o}_{t+1} = \theta(\vb{s}_{t+1})$ and a reward $\vb{r}_{t+1} = r(\vb{s}_t, \vb{a}_t)$. 
\textbf{Causal POMDPs} explicitly model the effects of causal factors on the transition and emission functions by dividing the state into the controllable state $\vb{s}^c_{t}$ and the causal factor, $\vb{\mathcal{H}}$. The causal factors of an environment cannot be manipulated by the agent, but their values affect the outcome of an action taken by the agent. Thus the transition function of the controllable state is:
\begin{equation}
\label{eq:transition}
\phi(\vb{s}^c_{t+1}| \vb{s}^c_{t}, f_{sel}(\vb{\mathcal{H}}, \vb{s}^c_{t}, \vb{a}_{t}),\vb{a}_{t})    
\end{equation}
where $f_{sel}$ is the implicit Causal Selector Function which selects the subset of causal factors affecting the transition defined as:
\begin{equation}
\label{eq:selector}
f_{sel}:\mathcal{H} \times \mathcal{S} \times \mathcal{A} \to \powerset(\mathcal{H})
\end{equation}
where $\powerset(\mathcal{H})$ is power-set of $\mathcal{H}$ and $f_{sel}(\mathcal{H}, \vb{s}^c_{t}, \vb{a}_{t}) \subset \mathcal{H}$ is the set of effective causal factors for the transition $\vb{s}_t \to \vb{s}_{t+1}$ i.e., $\forall v\neq v'$ and $\forall\vb{h}_j\in f_{sel}(\mathcal{H}, \vb{s}^c_{t}, \vb{a}_{t})$: 
\begin{equation}
  \phi(\vb{s}^c_{t+1}|do(\vb{h}_j = v), \vb{s}^c_{t}, \vb{a}_{t}) \neq \phi(\vb{s}^c_{t+1}|do(\vb{h}_j = v'), \vb{s}^c_{t}, \vb{a}_{t})
\end{equation}

where $do(\vb{h}_j)$ corresponds to an external intervention on the factor $\vb{h}_j$ in an environment. 

Intuitively, this means that if an agent takes an action $\vb{a}_t$ in the controllable state $\vb{s}^c_t$, the transition to $\vb{s}^c_{t+1}$ is caused by a subset of the causal factors $f_{sel}(\mathcal{H}, \vb{s}^c_{t}, \vb{a}_{t})$. For example, if a body on the ground (i.e., state $\vb{s}^c_t$) is thrown upwards (i.e., action $\vb{a}_t$), the outcome $\vb{s}_{t+1}$ is caused by the causal factor gravity (i.e., $f_{sel}(\mathcal{H}, \vb{s}^c_{t}, \vb{a}_{t})=\{\text{gravity}\}$), a singleton subset of the global set of causal factors. The $do()$ notation expresses this causation. If an external intervention on a causal factor is performed, e.g., if somehow the value of gravity was changed from $v$ to $v'$, the outcome of throwing the body up from the ground, $\vb{s}_{t+1}$, would be different. 
\subsection{Algorithmic Information Theoretic View on Causality}
\begin{figure*}
    \centering
    \includegraphics[width=\textwidth, height=140pt]{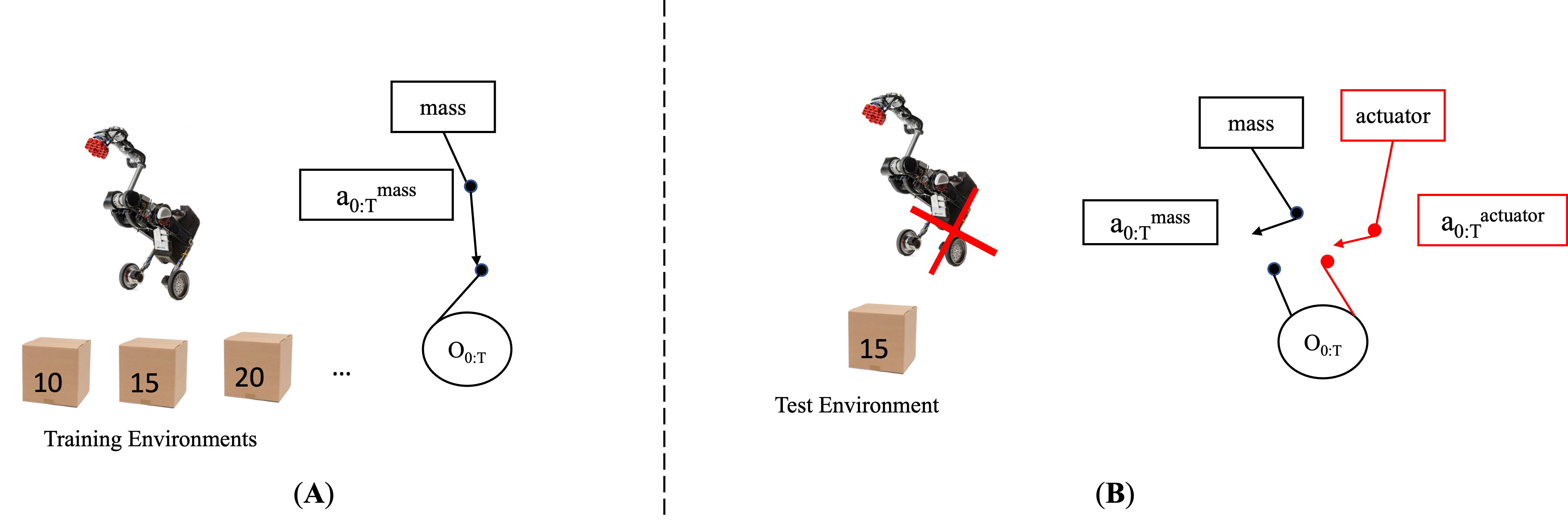}
    \caption{\textbf{Out-of-Task Distribution Transfer.} Pane \textbf{A} shows a training-time scenario where an agent learns to interact with environments containing objects for varying values of mass. The causal graph is gated as particular action sequences either obfuscate or underscore the effects of certain causal factors. Pane \textbf{B} represents the inference-time scenario where the a causal factor, actuator health, held constant during training is varied at inference.}
    \label{fig:gatedGraph}
\end{figure*}

Causality can be motivated from the perspective of algorithmic information theory \citep{JanSch10}. Consider the Gated Directed Acyclic Graph of the observed variable $\vb{O}$ and its causal parents (\Cref{fig:gatedGraph}). Each causal factor has its own causal mechanism, jointly bringing about $\vb{O}$. The action sequence $\vb{a}_{0:T}$ serves a gating mechanism, allowing or blocking particular edges of the causal graph using the implicit Causal Selector Function (\Cref{eq:selector}). A central assumption of our approach is that causal factors are independent, i.e., the Independent Mechanisms Assumption \citep{Schoelkopf2012, parascandolo2018learning, scholkopf2019causality}. The information in $\vb{O}$ is then the sum of information “injected” into it from the multiple causes, since, loosely speaking, for information to cancel, the mechanisms would need to be algorithmically dependent \citep{JanSch10}. Thus, the information content in $\vb{O}$ will be greater for a larger number of independent causal parents in the graph. 
\begin{equation}
\label{eq:informationProp}
L(\vb{O}) \propto |PA(\vb{O)})|
\end{equation}
where $L(\cdot)$ is the Minimum Description Length (MDL), a tractable substitute of the Kolmogorov Complexity \citep{rissanen1978modeling, grunwald2004tutorial}). 
\subsection{Causal Curiosity}
Causal curiosity \citep{sontakke2021causal} allows an RL agent to discover sequences of actions that bring out the effect of a single causal factor while ignoring the effects of all other. This is similar to how a human scientist studying multiple mechanisms in their environment would behave whilst following the One-Factor-at-a-Time (OFAT) paradigm of experiment design \citep{fisher1936design}. For e.g., when interacting with objects of varying mass and shape, a human scientist will learn a perfect lifting sequence that grasps all shapes and then use it to test out the mass of each object. 

Thus, \emph{Causal Curiosity selects one among multiple competing causal mechanisms} and generates a sequence of actions that bring out the effect of the selected mechanism. This is done by attempting to learn a simple model of the environment with capacity low enough to learn about only a single causal mechanism at a time. One could conceive of this by assuming that the generative model for $\vb{O}$, $\vb{M}$ has low Kolmogorov Complexity. A low capacity bi-modal model is assumed. The Minimum Description Length (MDL), $L(\cdot)$  is utilized as a tractable substitute of the Kolmogorov Complexity \cite{rissanen1978modeling, grunwald2004tutorial}).  Subsequently, the following optimization problem is solved.  
\begin{equation}
\label{eq:new_mdl}
\vb{a}_{0:T}^* = \argmin_{\vb{a}_{0:T}}({L(\vb{M}) + L(\vb{O}|\vb{M})})
\end{equation}
where each observed trajectory $\vb{O}=\vb{O}(\vb{a}_{0:T})$ is a function of the action sequence. Thus the resulting action sequence from the optimization in \Cref{eq:new_mdl} will result in an action sequence that brings out the effect of a single caual factor. Having established this, we now introduce a causal perspective on transfer.

\subsection{Causal Perspective on Transfer}
Consider the set of POMDPs $P=\{\vb{p}_0,\vb{p}_1, \ldots\}$ parameterized by the tuple ($\mathcal{O}$, $\mathcal{S}$, $\mathcal{A}$, $\phi$, $\theta$, r, $H'\subset H$) with observation space $\mathcal{O}$, state space $\mathcal{S}$, action space $\mathcal{A}$, the transition function $\phi$, emission function $\theta$, and the reward function $r$, with the set of causal factors $H'\subset H$, i.e., subset of the global causal factors, varied over a range of values and the remaining $H-H'$ held constant.
\begin{definition}[In-Task-Distribution Transfer]
  \label{def:ID_Transfer}
  An in-task-distribution transfer occurs when an agent trained on $P$ is launched into a POMDP $\vb{p}'$ where $\forall~\vb{h} \in H-H'$ the values assumed by $\vb{h}$ remain unchanged (assume the same values as in P). 
\end{definition}

\begin{definition}[Out-of-Task-Distribution Transfer]
  \label{def:OOD_Transfer}
  An Out-of-Task-distribution transfer occurs when an agent trained on $P$ is launched into a POMDP $\vb{p}'$ where $\exists~\vb{h} \in H-H'$ which assumes a value different from the value it had in $P$. 

\end{definition}
Consider a transfer learning agent training to lift cubes with varying masses and sizes, i.e., $H' = \{mass, size\}$. An In-Task-Distribution Transfer scenario occurs if at test-time it encounters an cube of an unseen mass/size combination. An Out-of-Task-Distribution scenario occurs if it is required to lift a cube with a broken actuator. This is because the causal factor $actuator \in H-H'$ which was held constant during training (agent trained is using a healthy actuator) is required to lift using a broken actuator at test-time. We would like to be able to detect such faults while generalizing to known causal factors.  

\section{Method}
\paragraph{Setup}
We consider the scenario where a learning agent is trained on a set of POMDPs $P=\{\vb{p}_0,\vb{p}_1, \ldots\}$ parameterized by the tuple ($\mathcal{O}$, $\mathcal{S}$, $\mathcal{A}$, $\phi$, $\theta$, r, $H'\subset H$) with observation space $\mathcal{O}$, state space $\mathcal{S}$, action space $\mathcal{A}$, the transition function $\phi$, emission function $\theta$, and the reward function $r$, with the set of causal factors $H'\subset H$, i.e., subset of the global causal factors, varied over a range of values and the remaining $H-H'$ held constant. 

We assume that the learning agent is able to learn $\mathbf{z}=Z_\phi(\mathbf{p})$, called \emph{belief function}, using each of the training environments which generates a representation for the intervened causal factors, i.e., $H'\subset H$. This assumption is quite general - the RL systems that are capable of performing well over different environments can be assumed to either explicitly model such representations (as in \citep{rakelly2019efficient, zintgraf2019varibad, perez2020generalized}) or implicitly (as in \citep{finn2017model, nagabandi2018learning}).
At test time, the agent is launched into a novel environment $\vb{p}'$ which is either an In-Task-Distribution Transfer (see \Cref{def:ID_Transfer}) or Out-of-Task-Distribution Transfer (see \Cref{def:OOD_Transfer}).
\subsection{Construction of the Belief Set}
The agent performs inference in the novel test environment $\vb{p}'$ using $Z_\phi(\mathbf{p}')$. We assume that the agent has access to $\{Z_\phi(\mathbf{p}): \mathbf{p}\in P\}$, i.e., the belief representation for the training environments. The agent then collects all training environments that lie near $\vb{p}'$ in the space of the learned belief functions into the ball $\mathcal{B}$ called the \emph{belief set} defined as,
\begin{equation}
    \mathcal{B}:=\{\vb{p}_i : d(q_{\phi}(\vb{z}|\vb{p}')||q_{\phi}(\vb{z}|\vb{p}_i)) < \epsilon\}
\end{equation}
where $d(\cdot|\cdot)$ is a distance function (e.g., Euclidean) in the latent space and $\epsilon$ is a design hyperparameter.

Thus, for example, in a lifting task of cubes of various masses, if the agent fails to lift a cube at test time, it constructs the belief ball consisting of the training environments with close representations,  i.e., heavy cubes and adds them to the belief set $\mathcal{B}$. Depending on the cause for the failure of the agent in lifting the cube, the situation goes into one of the following branches: (1) The test environment requires an \textbf{In--Task-Distribution Transfer}, i.e., the test-time block is actually a heavy block or (2) The test environment requires \textbf{Out-of-Task-Distribution Transfer}, i.e., a broken actuator makes a light block seem heavy.  
\subsection{Belief Verification}
Subsequently, the agent optimizes causal curiosity on $\mathcal{B} \cup \{\vb{p}'\}$. As in \Cref{eq:new_mdl}, a low capacity binary clustering model is considered. Thus, the following optimization procedure is implemented:
\begin{equation}
\label{eq:implementation_distance_cost}
\begin{split}
  \argmax_{\vb{a}_{0:T} \in \mathcal{A}^T}[\min \{d(\vb{o}_{0:T}, \vb{o'}_{0:T}): \vb{o}_{0:T}\in C_1, \vb{o'}_{0:T}\in C_2\} - \\ \max\{d(\vb{o}_{0:T}, \vb{o''}_{0:T}): \vb{o''}_{0:T}, \vb{o}_{0:T}\in C_1\} - \\
  \max\{d(\vb{o'}_{0:T}, \vb{o'''}_{0:T}): \vb{o'}_{0:T}, \vb{o'''}_{0:T}\in C_2\}] \end{split}
\end{equation}
where $\vb{O}$ is the observation obtained by applying action sequence $\vb{a}_{0:T}$. Clusters $C_1~\text{and}~C_2$ represent the bimodal model. 

\begin{figure}
    \centering
    \includegraphics[height=80pt]{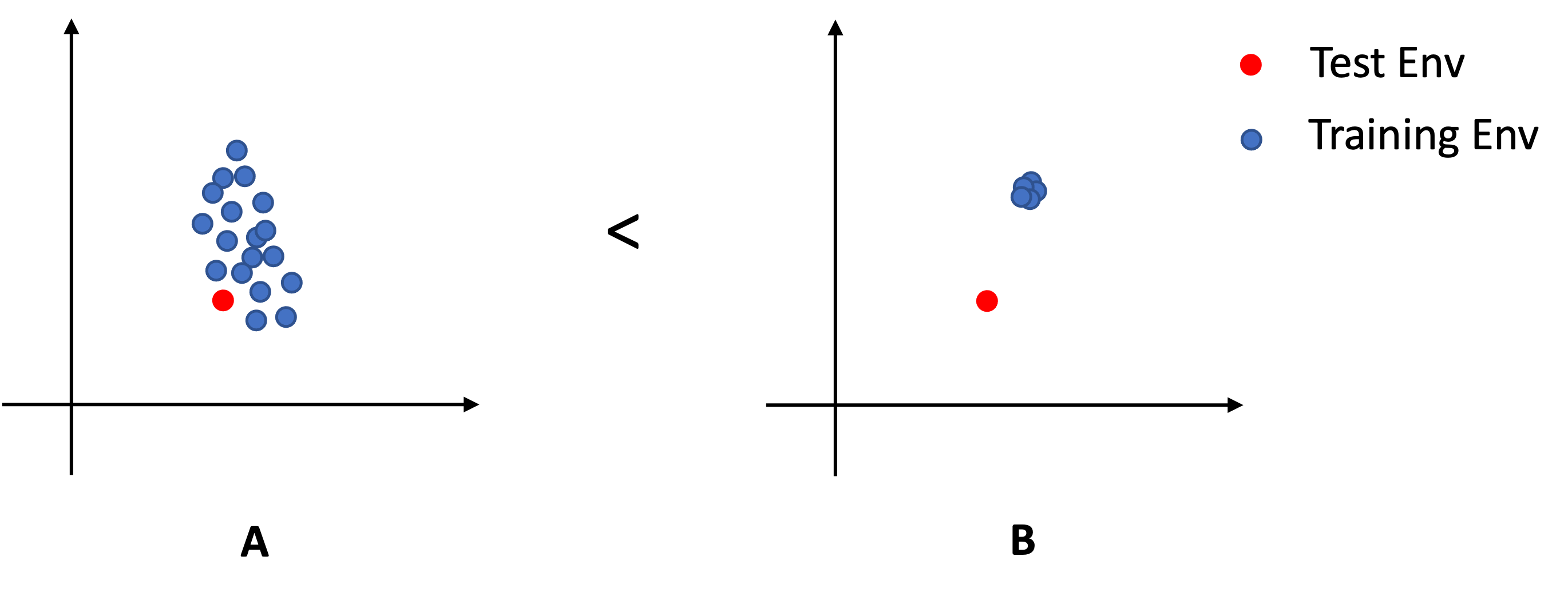}
    \caption{\textbf{Visualization of the Observation $\vb{O}$.} Pane \textbf{A} represents the observation variables obtained after the optimization of \Cref{eq:implementation_distance_cost} during an In-Task-Distribution Transfer. Causal Curiosity will be quite low in such a case as the bi-modal clustering would be poor. Pane \textbf{B} represents the case when OOTD transfer occurs - the causal curiosity reward would be high as the bi-modal clustering would be near-perfect.}
    \label{fig:CCVis}
\end{figure}

\paragraph{In-Task Distribution} If the test environment $\vb{p}'$ is In-Task Distribution, then the variance of values assumed by the causal factors $H'$ in the set of environments $\mathcal{B} \cup \{\vb{p}'\}$ is small and the clusters are \emph{not well-separated}. Thus optimizing causal curiosity as in \Cref{eq:implementation_distance_cost} will produce action sequences that result in observations that cluster in a distributed manner as in pane \textbf{A} of \Cref{fig:CCVis}.

Intuitively, if the agent has learnt to interact with blocks of various masses and at test time is presented with a heavy block, the outcome of its interaction with the test block (i.e., $\vb{p}'$) will not differ significantly in comparison with the heavy blocks it interacted with during training.
\paragraph{Out-of-Task-Distribution} However, during the optimization of \Cref{eq:implementation_distance_cost} in the OOTD case, 2 competing causal mechanisms will exist - one induced by the set $H'$ and the other from the set $H-H'$. The mechanism caused by $H-H'$ will however dominate as all environments in $\mathcal{B}$ will have the same values for $H-H'$ while $\vb{p}'$ will have a different value. Thus, the resulting clusters for the causal mechanism from $H-H'$ will be \emph{well-separated}. Subsequently, the causal curiosity reward (\Cref{eq:implementation_distance_cost}) will be higher for selecting the causal mechanism induced by $H-H'$.

Intuitively, as in the above example of an agent interacting with blocks of varying masses but constant size $\xi$, if at test time, the agent is provided with a block of low mass and a new size $\xi'\neq \xi$, the causal curiosity reward for the size mechanism will be higher because a perfect binary clustering is possible (as in pane \textbf{B} in \Cref{fig:CCVis}) where one cluster contains observations from training environments (\textcolor{blue}{blue cluster}) corresponding to size $s$ while the other cluster corresponds to the test environment with size $s'$ (\textcolor{red}{red cluster}).   
Thus, if the test environment $\vb{p}'$ lies in its own cluster after optimizing causal curiosity on $\mathcal{B} \cup \{\vb{p}'\}$, then  \approachName~concludes $\vb{p}'$ to be OOTD, i.e., 
\[
   \textrm{Is\_OOTD}(p')= 
\begin{cases}
    1,& \text{if } \vb{p}'~\text{lies in its own cluster}\\
    0,              & \text{otherwise}
\end{cases}
\]
Note, the causal curiosity for a known causal factor, (In-Task-Distribution Transfer) will be less than the causal curiosity for an unknown causal factor (OOTD Transfer) as seen in \Cref{fig:CCVis}. 

\subsection{Probabilistic Baselines}
A natural extension of Model-based learning methods for OOTD is possible. We question whether such an extension yields good results. We test whether OOTD Detection is possible by simply learning a model of the training and test environments and using using the discrepancy of the outputs to detect whether the learnt test-time model represents a task from OOTD.  

We utilize an ensemble of probabilistic neural networks (PNNs) \citep{lakshminarayanan2017pne, Chua2018PDM}, which is a generative neural network whose output neurons parameterize a probability distribution $p_\theta(y|\vb{x})$; a mean value corresponds to the believed label $\hat{y}$ along with some degree of uncertainty $\theta$. 

For an environment $\vb{p}$, we estimate the environment transition function $\vb{\phi}_{\vb{p}}(\vb{s}_{t+1} | \vb{s}_t, \vb{a}_t)$ using an ensemble of PNNs $f^{\vb{p}}_{\phi}(\vb{s}_{t+1} | \vb{s}_t, \vb{a}_t))$. 
We are interested in the disagreement between a novel test environment \testEnv relative to a training environment \trainEnv{} which we measure using the relative entropy between $f^{\vb{p'}}_{\phi}$ and $f^{\vb{p}}_{\phi}$ given by $KL(f^{\vb{p'}}_{\phi} || f^{\vb{p}}_{\phi})$:
\begin{equation}
\label{eq:KL_div}
\begin{split}
    D_{KL}(f^{\vb{p'}}_{\phi}||f^{\vb{p}}_{\phi}) = &\frac{1}{2} [\log{\frac{ |\Sigma(f^{\vb{p}}_{\phi})|} {|\Sigma(f^{\vb{p'}}_{\phi})|}} 
    - k + \\ (\mu(f^{\vb{p'}}_{\phi}) - &\mu(f^{\vb{p'}}_{\phi}))^T \Sigma^{-1}(f^{\vb{p}}_{\phi}) (\mu(f^{\vb{p'}}_{\phi}) - \mu(f^{\vb{p}}_{\phi})) + \\ \Tr{\Sigma^{-1}(f^{\vb{p}}_{\phi}) \Sigma(f^{\vb{p'}}_{\phi})}]
\end{split}
\end{equation}
where $k$ is the dimensionality of the environment's observation space; $\vb{s}_t \in \mathbb{R}^k$, $\Sigma(\cdot)$ is covariance and $\mu(\cdot)$ is mean and $\Tr(\cdot)$ is trace.

We use Negative Log Loss as a scoring rule for PNNs, and KL divergence as a measure of distribution disagreement. We utilize a thresholding technique here to detect OOTD. We train  $f^{\vb{p'}}_{\phi}$ and $f^{\vb{p}}_{\phi}$ over multiple random seeds and use the mean $KL(f^{\vb{p'}}_{\phi} || f^{\vb{p}}_{\phi})$ of the first $k$ seeds as the threshold.  
Detection is then performed using:
\[
    \textrm{Is\_OOTD}(p')= 
\begin{cases}
    1,& \text{if } KL(f^{\vb{p'}}_{\phi} || f^{\vb{p}}_{\phi}) >  threshold\\
    0,              & \text{otherwise}
\end{cases}
\]

\section{Experiments}
Our work has 2 main thrusts - the discovered \emph{experimental behaviors} and the \emph{Out-of-Task-Distribution Detection} obtained from the outcome of the behaviors in environments. We visualize these learnt behaviors and verify that they are indeed semantically meaningful and interpretable. We quantify the utility of the learnt behaviors to perform OOTD detection. 
\textbf{Causal World.} We use the Causal World Simulation \citep{ahmed2020causalworld} based on the Pybullet Physics engine to test our approach. The simulator consists of a 3-fingered robot, with 3 joints on each finger. We constrain each environment to consist of a single object that the agent can interact with. The causal factors that we manipulate for each of the objects are size and mass of the blocks and the damping factor and control frequency of the robotic motors. The simulator allows us to capture and track the positions and velocities of each of the movable objects in an environment. 
\newline\textbf{Mujoco Control Suite.} We also perform OOTD Detection on 3 articulated agents that try to learn locomotion - Half-Cheetah, Hopper, and Walker. For each agent type, the causal factors that we intervene on include the mass of the robot, and wind and gravity in the environment, and the friction between the robot actuators and the ground.

\subsection{Generalized Experimental Setup}
To test our approach, we train a transfer RL algorithm - in our case, Causal Curiosity \citep{sontakke2021causal} on multiple environments with causal factor $A$ assuming values $A=a\in\mathbbm{A}$ where $\mathbbm{A}$ is a set of values causal factor $A$ can assume. For example, we train an agent to interact with blocks of varying masses (here mass is causal factor $A$). At test time, we generate a range of values of a causal factor $B$ previously held fixed. Thus, in the above example, having been trained on varying values of mass, we now generate a range of values for the control frequency (causal factor $B$) of the robot actuators (previously held constant at some $\eta$). For each pair of values of $(B, A)$ causal factors, we report the accuracy of detection over 10 random seed experiments. Thus, in the above example, all environments with $(control=\eta,mass=m)$ are considered In-Task-Distribution Transfer while all others are Out-of-Task Distribution.  
\newline\textbf{Interpreting Results}
For each pair of test-time and training-time causal factors, we vary both over a range of values. Consider for example \Cref{fig:CW_rest}, where we vary the mass of the blocks the robot interacts with during training and at test-time, it receives environments with a different perception frequency. During training in \Cref{fig:CW_rest} Pane \textbf{C}, the perception frequency was at $1$ (corresponding to the column at Perception Frequency = $1$). For each pair of (Control Frequency, Mass), we run our method over 10 random seeds. The value at each $(x,y)$ position corresponds to the number of times during the 10 runs, \approachName~concluded that the test environment was OOTD. \Cref{fig:CW_rest} Pane \textbf{C} is an example of perfect detection - no false positives (column above $x=1$ is zero) and $100\%$ detection when Perception Frequency is varied. Other experiments depict varying degrees of detection success.
\subsection{Causal World Experiments}

During training, we vary either the mass or size of the block in an environment. At test time, the agent interacts with an environment with 3 possible errors - \textbf{(1) Distributional Shift of the Environment}: Change in the physical features of the block (2) \textbf{Perception Defect}: Frequency of perception changes (i.e., framerate of sensors) which affects the perception-to-action loop and \textbf{(3) Actuator Defect}: the damping coefficient of the arm actuators is varied, which affects the dynamics of the robotic arm.  

\begin{figure*}
    \centering
    \includegraphics[width=\textwidth, height=200pt]{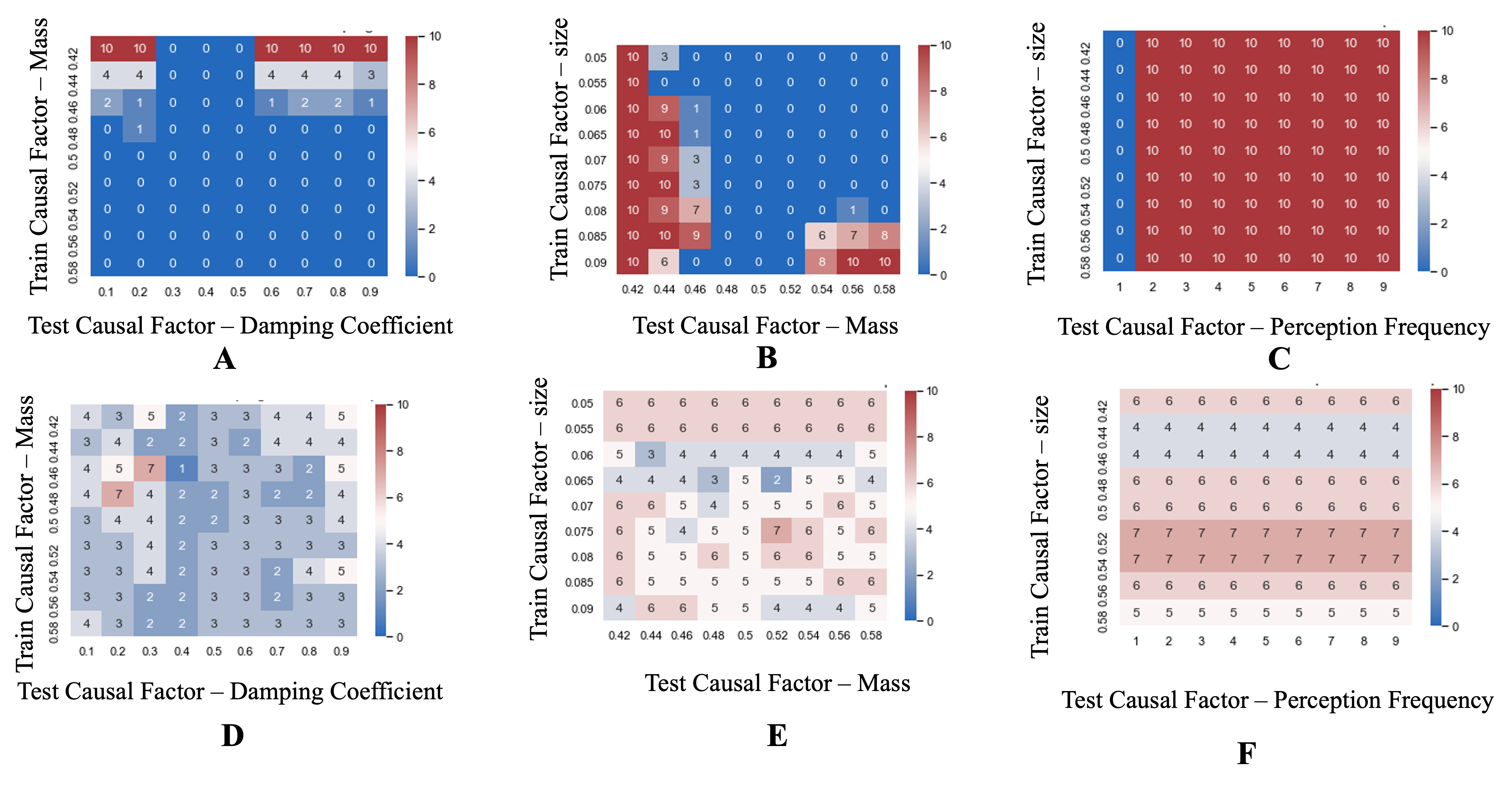}
    \caption{\textbf{Causal World experiments.} Subfigures \textbf{A} -- \textbf{C} refer to \approachName, and subfigures \textbf{D} -- \textbf{F} refer to the probabilistic baseline. Each $(x, y)$ pair on the plot corresponds to an $(unseen, seen)$ pair of causal factors. The value at each $(x,y)$ pair depicts the performance of each method across measure performance as the number of correctly classified environments on 10 different random seeds. In-distribution value of mass is $0.5$; Damping Coefficient is $0.5$; Perception frequency is $1.0$. Ideally, the column above the training value of the OOTD causal factor should be 0, while all other columns should be 10 as is the case in Pane \textbf{C}. }
    \label{fig:CW_rest}
\end{figure*}

\Cref{fig:CW_rest} depicts experiments with mass and size of blocks as training-time causal factors and Perception Frequency, Damping Factor and Mass as test-time causal factors. Each of these experiments yield no false positives as the columns above in the fixed value of the test-time causal factor have zero detections. \Cref{fig:CW_rest} Pane \textbf{C} shows the agent has a perfect detection performance in detecting the perception defect. \Cref{fig:CW_rest}  panes \textbf{A} and \textbf{B} show that detection is successful when the test-time value of the unseen causal factor is some distance away from its constant value during training. The agent is more likely to detect an unknown causal factor when we make a larger change to it (larger values near the left and right border). 

\subsection{Mujoco Experiments}
\begin{figure*}[h]
    \centering
    \includegraphics[width=\textwidth, height=600pt]{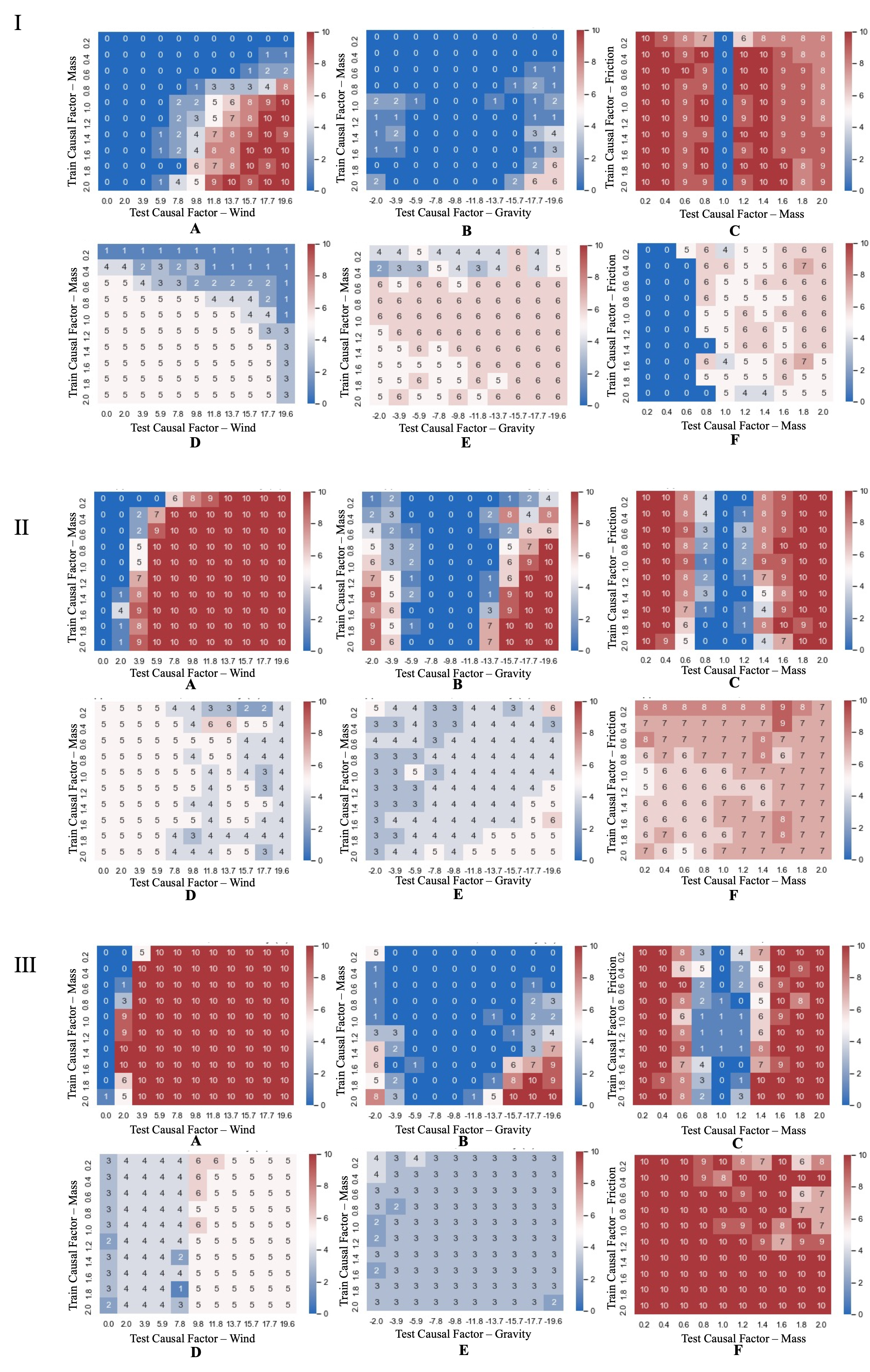}
    \caption{\textbf{Mujoco experiments.} Plots \textbf{I, II} and \textbf{III} correspond to Hopper, Walker and Cheetah environments respectively. Within each, subfigures \textbf{A} -- \textbf{C} refer to \approachName, and subfigures \textbf{D} -- \textbf{F} refer to the probabilistic baseline. Each $(x, y)$ pair on the plot corresponds to an $(unseen, seen)$ pair of causal factors. The value at each $(x,y)$ pair depicts the performance of each method across measure performance as the number of correctly classified environments on 10 different random seeds. In-distribution value of wind is $0.0$; gravity is $-9.8$; mass is $1.0$.}
    \label{fig:cheetah}
\end{figure*}

We perform experiments with the Mujoco control suite \citep{todorov2012mujoco} as well. During training, we manipulate the mass of the friction and the friction coefficients between the agent actuators and ground. At test time, we manipulate the wind and gravity in the environment and the mass of the agent. The in-distribution value of wind is $0.0$, gravity is $-9.8$ and mass is $1.0$. Discerning wind while being invariant to agent mass (Panes \textbf{A} and \textbf{D} in each sub-figure of \Cref{fig:cheetah}) is a relatively easy endeavour with the half-cheetah resulting in the highest accuracy across each of the random seeds. The task of discerning mass while being invariant to friction also yields high accuracy of detection when the  test mass  varies significantly in comparison with training time mass (red columns on right and left edges of Panes \textbf{C} and \textbf{F} in \Cref{fig:cheetah}). However, it suffers from poor detection at $0.8\times$ and $1.2\times$ the default mass for cheetah and hopper. The hardest task is that of discerning gravity while being invariant to mass - a task that requires discovering the \textbf{Galilean Equivalence Principle}, i.e., that the acceleration due to gravity is independent of mass. While the success of \approachName~is limited when  gravity is in the vicinity of $9.8$, it begins to successfully learn to detect changes in gravity as it deviates from $9.8$.    



\subsection{Interpretation of Learned Behaviours}
For visualizations of our method, see \href{https://galil-ai.github.io/}{\textbf{here.}} We analyze whether the discovered experimental behaviors are actually semantically meaningful. We find that the agent is able to discover many semantically meaningful behaviors that underscore the effect of a new causal factor previously held constant during training. Chiefly, we find that the 17th century philosopher Galileo Galilei and his namesake \approachName~agree that mass and gravitational acceleration are decoupled - \approachName~learns a free-falling behavior that mimics Galileo's experiments of dropping objects to discern the gravity of an environment whilst being invariant to the agent mass. 

In Mujoco Experiments, for mass as training time causal factor and wind as test-time factor, the agent learnt to use its body as a sail and allow the wind to carry it along. It also learnt to do front-flips and rolls in the direction of the wind, using the wind to help it along. For friction as training causal factor and mass as the test time causal factor, the agent also learnt to perform headstands to test out its mass while avoiding any horizontal locomotion allowing it to be invariant to the friction coefficients. 

In Causal World Experiments, for size as training time causal factor and mass as test-time factor, the agent learnt a relay-kick action when one of the finger push the object to the other finger, who makes a further push on it. This relay can only be finished on small mass blocks, thus distinguishing the causal factor.

\section{Conclusion}
In this work, we propose a novel task - that of Out-of-Task Distribution (OOTD) Detection and offer a causally inspired solution for the same. We find that simplistic extensions of existing model-based methods result in suboptimal performance with either low-detection accuracy and high false positive rate. We show the efficacy of our method in both a variety of embodied robotic environments spanning 2 simulation engines. We find \approachName~has the ability learn complex causal mechanisms and is a first step towards safer transfer/meta-RL. 
\clearpage
\bibliographystyle{plainnat}
\bibliography{sample}
\end{document}


%

%

\onecolumn
\aistatstitle{Instructions for Paper Submissions to AISTATS 2022: \\
Supplementary Materials}

\section{Implementation details}

\subsection{Planner}
The Experiment Planner consists of a uniform distribution control planner with Cross Entropy Method Model Predictive Control. Each planner is initialized to a uniform distribution $\mathcal{U}(controlLow, controlHigh)$. 
For \textbf{Mujoco} experiments, each planner consists of 20 sampled plans per iteration. Each sampled plan consists of 6 control signals applied for a duration of 10 frames, for a total of 60 frames per episode. 
For \textbf{Causal World} experiments, each planner similarly consists of 20 sampled plans per iteration, with each action applied for a longer duration, for a total of 198 frames per episode.
In both cases, each sampled plan is applied to each of the considered environments. At the end of each training iteration, the top $10\%$ of plans are used to update the agent's action distribution. In total, training required 20 full iterations.

In general, 
during training, the agent learns a sequence of actions to maximize the Causal Curiosity reward across 9 different environments, e.g. block mass of 0.1 to 0.9 with step 0.1. The learned action sequence will group the training environments into 2 clusters, such as a large mass cluster and a small mass cluster. Then, using the action sequence which maximizes the desired optimization problem, the agent is tested in an OOTD environment and classifies said environment to one of its prior two belief clusters according to some distance function. 
Following the creation of the agent's belief cluster (cluster containing test environment), we then conduct the same training procedure again on this new environment with its belief cluster environments. If the new clustering result will separate the test environment in its own cluster, while others remain in the other one, we say the agent made a detection of the unknown causal factor. We run such experiments 10 times over different random seeds on different training-test environment pairs covering various unknown causal factors. To prevent over-fitting on In-Distribution tasks, training is performed on slightly different values of causal factors than what is seen during testing, e.g.~train on $mass = 0.24m$, test on $mass = 0.20m$.

\subsection{Modifying Environments}

\textbf{Mujoco.} For mass experiments, we vary the normal mass of the robot $(m)$from $0.2m$ to $2m$. Similarly when modifying friction values in the environment, we change the friction coefficient $\eta$ between the robot's actuators and the ground from $0.2\eta$ to $2\eta$. For gravity experiments, we modify the absolute value. The ground truth $(g_z = -9.81)$ from low gravity $g_z=-2.0$ to high gravity $g_z=-19.6$ for a total of 10 values. In wind experiments, we deviate from the typical value of $0.0$ (no wind) for $10$ values between $2.0$ to $19.6$.
In \textbf{Causal World}, we are able to modify the absolute mass and shape of the block the agent interacts with. Changing the perception value of the robot is equivalent to modifying the skip-frame value of the robot's controller. Larger values of skip-frame leads to a slower refresh rate of the robot's sensors, and leads to less controllable actions. 

\section{Probabilistic Baseline}




To evaluate our prediction model, we design a baseline solely based on the first round of training. If the posterior distribution $\phi(s_t+1|s_t,a_t)$ learned in the test environment is more than a reasonable large threshold distance away than the distribution learned in the training environments, as measured by KL divergence, we denote it as a detection. 

We assume for a Causal POMDP $\vb{p}$, the agent's observations at timestep $t$ is a random variable generated from a Gaussian distribution, such that $\vb{s}_{t+1} \sim \mathcal{N}(\mu_{s+t}, \Sigma)$.
For each  $(unseen, seen )$ pair of causal factors, we train an ensemble of probabilistic neural networks, each which output a mean vector $\mu_{\vb{p}}$ and a diagonal covariance matrix $\Sigma_{\vb{p}}$. 
Each ensemble is a uniformly-weighted mixture model, and we combine the predictions as $p(y|\vb{x}) = M^{-1} \sum_{m=1}^M p_{\theta_m} (y | \vb{x}, \theta_m)$. The prediction is then a mixture of Gaussian distributions. We assume the covariance matrix $\Sigma_{\vb{p}}$ is a diagonal matrix. For ease of computation, we further approximate the ensemble prediction as a Gaussian whose mean and variance are respectively that of the mixture; $\mu^* = M^{-1}\sum_{m=1}^{M}\mu_m$ and $\sigma^* = M^{-1} \sum_m(\sigma_m^2 + \mu_m^2) -\mu^{*2}$.

As training data for each network, we use the set of all $(state, action)$ pairs gathered during the first round of training our algorithm; $\mathcal{D} = \{(\vb{s}_t, \vb{a}_t), \vb{s}_{t+1} \}_{t=0}^T$.


In practice, each network was trained for 40 epochs across 10 random weight initializations, with a learning rate of $0.001$ and Adam as the optimizer. We used an ensemble size of $M = 10$ for each experiment.
To set the threshold, we gathered training data from 5 seeds unseen by our method, for every $(unseen, seen)$ pair of causal factors, and took the average value of the KL divergence of the test environment with respect to the training environments. 

The ensemble model was inspired in part due to the observation that different random weight initialization produced different distribution predictions from one another. However, ultimately we remark that the overall performance of the baseline did not differ significantly if an ensemble was not used.





\section{Error Analysis}

As is evident our result figures, agents are more likely to detect an unknown causal factor when a larger change is made to its value (larger values further away from the in-distribution value column).
Agents are less likely to detect a change to their environment when the percentage change of the training causal factor in its belief cluster is large while the percentage change of the unknown causal factor is small. 
In Causal World, we found different factors to have different significance levels. In general, $Framerate >> Size > Damping > Mass >> Friction$. In an environment setting, the agent is able to detect a causal factor if the training factor has a lower significance value than the causal factor.
For example, after examining the visualizations, we find that when the test environment is clustered together with heavy masses, the heavy mass dominates the effect of the damping, and the agent learns to further separate heavy blocks from light blocks in this new setting.

In another word, maximizing Causal Curiosity will separate the most significant factor (the significance is determined by the nature of the factor and the variance of it across all training environments) into 2 clusters. Each cluster will have a smaller variance of the training factor, thus lower significance. When we continue this process until the significance of the training factor is low enough in a cluster, the next significant factor (causal factor in our case) will be taken into consideration in the next training. 


In Mujoco, after examining the visualizations, we postulate that agents with high action and observation spaces, such as Walker, are more prone to confusing actions such as front-flips and rolls with being pushed by the wind. This could be due to frequent relative change in position from one of the robot's sensors to another. Agents with small action and observation spaces, such as the Hopper, suffer less from this sensor confusion because their observations rely more on their absolute position in the environment. In Mujoco, a robot's absolute position in their environment was one of the most important factors in determining whether an environment OOTD or not for many of the considered causal factors.

Finally, compared to the probabilistic baseline, we would like to point out our method shows a more anthropomorphic 
response to varying values of causal factors.
Consider the following example of how a human might see if its windy outside before leaving the house. 
A human may still need to check the weather report, or look at the leaves blowing in the wind, to determine if there is slight or no breeze outside. However, if there is a significant gust, one would simply be able to tell by sticking her arm out the window.
Similarly, our method shows a similar (lack-of) sensitivity to certain varying causal factors. 
On the other hand, the baselines do not show such sensitivity. The tendency to predict the same value across multiple $(unseen, seen)$ pairs is likely due to the data generative process used to gather the training data. Not all action sequences may bring forth the causal factor's influence in the environment, but we consider all action sequences generated by the planner during the initial training process. Our method on the other hand, only considers the best action sequence; the action sequence which maximizes the optimization problem discussed in the main text.


\clearpage
\bibliographystyle{plainnat}
\bibliography{sample}